\documentclass[letterpaper, 10 pt, conference]{ieeeconf}
\makeatletter
\let\MYcaption\@makecaption
\makeatother

\usepackage[font=footnotesize]{subcaption}

\makeatletter
\let\@makecaption\MYcaption
\makeatother

\usepackage{graphics} 
\usepackage{epsfig} 
\usepackage{amsmath} 
\usepackage[T1]{fontenc}
\usepackage{bm}
\usepackage{cite}
\usepackage{here}
\usepackage{booktabs}
\usepackage{threeparttable}
\usepackage{multirow}
\usepackage{color} 
\usepackage{cases}
\usepackage{siunitx}

\usepackage{algorithm}
\usepackage{algpseudocode}
\usepackage{makecell}

\usepackage{amssymb}

\usepackage{float}

\IEEEoverridecommandlockouts                              
\title{\LARGE \bf Multi-Modal Interaction Control of Ultrasound Scanning Robots with Safe Human Guidance and Contact Recovery}
\author{Xiangjie Yan, Yongpeng Jiang, Guokun Wu, Chen Chen, Gao Huang, and Xiang Li
\thanks{X. Yan, Y. Jiang, C. Chen, G. Huang, and X. Li are with Department of Automation, Tsinghua University. G. Wu is with Shenzhen International Graduate School, Tsinghua University.
This work was supported in part by the Science and Technology Innovation 2030-Key Project under Grant 2021ZD0201404, in part by the National Natural Science Foundation of China under Grant U21A20517 and 52075290, 
and in part by the Institute for Guo Qiang, Tsinghua University.  Corresponding author: Xiang Li (xiangli@tsinghua.edu.cn)}
}

\begin{document}
\maketitle

\begin{abstract}
Ultrasound scanning robots enable the automatic imaging of a patient's internal organs by maintaining close contact between the ultrasound probe and the patient's body during a scanning procedure.
Comprehensive, high-quality ultrasound scans are essential for providing the patient with an accurate diagnosis and effective treatment plan.
An ultrasound scanning robot usually works in a doctor-robot co-existing environment, hence both efficiency and safety during the collaboration should be considered. 
In this paper, we propose a novel multi-modal control scheme for ultrasound scanning robots, in which three interaction modes are integrated into a single control input. Specifically, the {\em scanning mode} drives the robot to track a time-varying trajectory on the patient's body under the desired impedance model; the {\em recovery mode} allows the robot to actively recontact the body whenever physical contact between the ultrasound probe and the patient's body is lost; the {\em human-guided mode} renders the robot passive such that the doctor can safely intervene to manually reposition the probe. 
The integration of multiple modes allows the doctor to intervene safely at any time during the task and also maximizes the robot's autonomous scanning ability. 
The performance of the robot is validated on a collaborative scanning task of a carotid artery examination.
\end{abstract}

\section{Introduction}
Ultrasound scanning is a common noninvasive health screening method in great demand\cite{shung2011diagnostic}. 
During an ultrasound scanning task, a doctor holds a scanning probe and moves it along a patient's body, maintaining close and stable contact to enable clear ultrasound imaging. Ultrasound scans can be incredibly labor-intensive tasks, with doctors typically performing several hours of imaging in a single day. Therefore,
deploying a robot to autonomously carry out these scanning tasks can help alleviate this problem. 

\begin{figure}[!tb]
\centering
\includegraphics[width=0.9\linewidth]{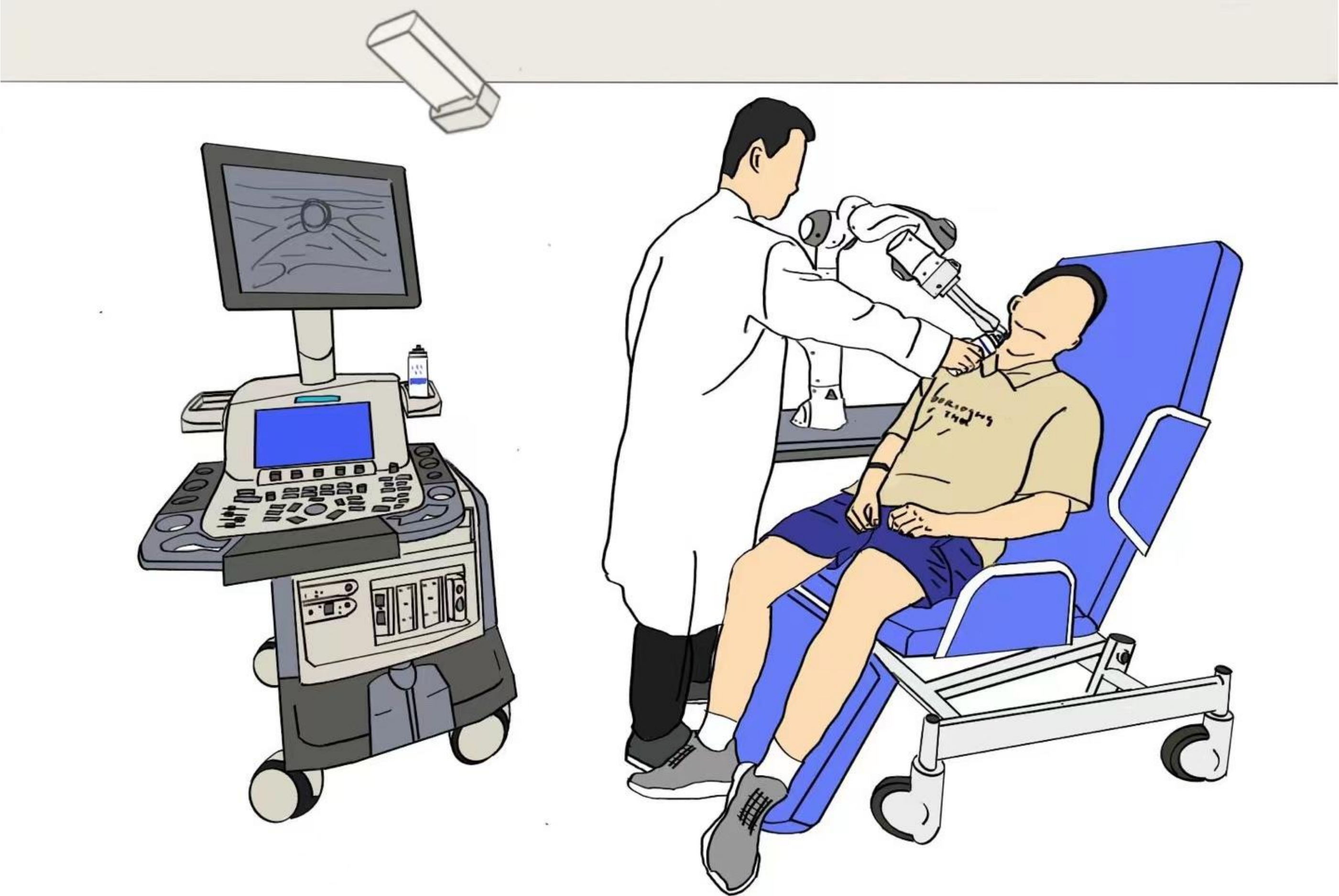}
\caption{An illustration of a carotid artery examination by an ultrasound scanning robot. The doctor is applying coupling gel and may directly manipulates the scanning probe to guide the motion of the robot. }\label{fig_scene}
\end{figure}

Since the scanning robot coexists with the doctor and also directly contacts the patient's body, it should guarantee their safety in two ways.
First, the contact force between the robot end effector, i.e., the probe, and the patient's body should be properly regulated. Too large a force could hurt the patient, however if the contact force is too small the scanning quality may be negatively affected. Second, the doctor may need to directly hold the robot end effector to correct its motion. In that case, the robot should become passive to follow his/her guidance. 
An illustration of a doctor-robot collaborative scanning task is shown in Fig.~\ref{fig_scene}.


In this paper, we proposes a new multi-modal interaction control scheme for ultrasound scanning robots to achieve both safety and scanning autonomy. Specifically, the proposed method allows the robot to safely carry out the scanning task autonomously while also allowing the doctor to safely intervene to lead the robot at any time. 
The multiple interaction modes are described as follows.
\begin{enumerate}
    \item[-] {\em Scanning mode}: the robot follows a trajectory on the patient's body under the desired impedance model to carry out the scanning task.
    
    \item[-] {\em Recovery mode}: whenever the contact is lost (e.g., due to a downward motion of the patient's body), the robot replans the reference trajectory to actively recontact the body and then continues the task.
    
    \item[-] {\em Human-guided mode}: when the end effector has moved too far away from the patient's body for self-recovery to be feasible, this mode is activated such that the robot becomes fully passive and waits for human guidance. This mode is also activated whenever the doctor takes hold of the robot end effector to intervene.
\end{enumerate}
The three interaction modes are smoothly integrated into a unified control input with a series of weighting factors. The transition between multiple modes is also embedded inside the control input and automatically achieved by the robot itself. Moreover, the stability of the closed-loop system is theoretically grounded, even during the transition between multiple modes. 
The proposed formulation combines the robot's ability and the doctor's knowledge/experience, making them complement each other. 
The performance of the proposed scheme is validated in real-world experiments involving cross-section carotid artery examinations.

\section{Related Works}
\subsection{Ultrasound Scanning Robot} 
Most ultrasound scanning robots can be categorized into three types: teleoperated ultrasound scanning robots, semi-autonomous ultrasound scanning robots, and fully autonomous ultrasound scanning robots \cite{Li2021}. In this paper, we mainly focus on the latter two types.

For a semi-autonomous ultrasound scanning robot, the operator and the robot have shared control \cite{selvaggio2021autonomy} over the movement of the scanning probe. A typical semi-autonomous procedure first involves a human operator manually moving the scanning probe to an approximate initial position. Then, using an image-based visual servoing method, the scanning probe tracks an existing feature in the ultrasonic image automatically \cite{nakadate2010implementation,hennersperger2016towards,kojcev2016dual,8002599}. 

In comparison, an autonomous ultrasound scanning robot possesses both high-level navigation/planning ability and low-level interaction control ability
\cite{Li2021}. The navigation systems can update the desired pose of the ultrasound probe online based on the data from sensors (such as force sensors, depth cameras, and the ultrasound images themselves) to obtain high-quality ultrasound images \cite{huang2018fully,welleweerd2021out,tan2022flexible}. Some works utilized reinforcement learning (RL) for navigation systems \cite{li2021image,bi2022vesnet} and achieved good performance at the price of low data efficiency.

While the aforementioned robots can perform the scanning task automatically, some of them cannot collaborate with a doctor or allow him/her to intervene during a scanning task. However, such a feature is important for an ultrasound scanning robot in a human-robot environment. For example, the doctor may
enter the robot's workspace to communicate with the patient, apply coupling gel, handle unexpected situations, correct or optimize the robot's motion, or provide diagnoses. In addition, allowing the doctor to intervene and lead the robot at any time during the scanning task increases patients' trust in and acceptance of the robot. 

\subsection{Interaction Control Scheme} 
For a robotic system physically interacting with the environment, it is important to design control schemes which ensure a safe interaction.
To enhance the interaction quality in task space, 
\cite{amanhoud2020force} proposed an RBF-based force correction model for dynamic compensation. In \cite{kronander2015passive}, a controller was proposed to ensure passivity in trajectory tracking tasks. In \cite{schindlbeck2015unified}, contact-loss stabilization was guaranteed by combining force tracking and impedance control.
These approaches primarily focused on flat or regular surfaces. In contrast, \cite{dyck2022impedance} was able to deal with complex arbitrary surfaces in task space.

In parallel, much progress has been achieved for a safe interaction in the null space of redundant robots
\cite{dietrich_overview_2015}. In \cite{sadeghian2011multi}, a null-space impedance control scheme was proposed to achieve multiple prioritized tasks at the acceleration level. The concept of extended task space was introduced and analyzed in \cite{oh1998extended,sadeghian2013task} to deal with the null-space interaction, which was accompanied by generalized force observers to eliminate the requirement of joint-torque measurement.

An ultrasound scanning robot working in a co-existing environment needs to be able to deal with interaction both in task space (i.e., between the robot end effector and the patient) and joint space (i.e., between the robot body and the doctor). In addition, the robot should be able to transition between multiple interaction modes to simultaneously enable autonomous operation and ensure safe intervention from doctors. The ability to do so is still an open issue for the interaction control methods discussed thus far.

\section{Preliminaries}
Consider the ultrasound scanning robot illustrated in Fig.~\ref{fig_scene}, where the scanning probe is installed on the robot end effector. During the scanning task, the doctor may frequently enter the robot's workspace and directly manipulate its end effector to guide its motion.  A 3D vision system is installed in the workspace for measurement.

The velocity of the robot end effector is given by 
\begin{equation}
\dot{\bm x}=\bm J(\bm q)\dot{\bm q},\label{velEqn}
\end{equation}
where $\bm x\hspace{-0.05cm}\in\hspace{-0.05cm}\Re^6$ denotes the position and the orientation of the robot end effector in Cartesian space, \(\dot{\bm x}\) is the velocity of the end effector, $\bm q\hspace{-0.05cm}\in\hspace{-0.05cm}\Re^n$ is the vector of joint angles, $n>6$ is the number of DOFs, and $\bm J(\bm q)\hspace{-0.05cm}\in\hspace{-0.05cm}\Re^{6\times n}$ is the Jacobian matrix from joint space to Cartesian space. Hence, this paper considers the redundant robot which provides much flexibility to adjust its end effector to suit the scanning task. Differentiating both sides of (\ref{velEqn}) with respect to time yields:
\begin{equation}
\ddot{\bm x} = \bm J(\bm q)\ddot{\bm q} + \dot{\bm{J}}(\bm q)\dot{\bm{q}}.\label{velEqn2}
\end{equation}

For the redundant robot, we have:
\begin{equation}
\bm{J}^{+}(\bm q)\triangleq\bm J^\mathsf{T}(\bm q)(\bm J(\bm q)\bm J^\mathsf{T}(\bm q))^{-1}\in\Re^{n\times 6},
\end{equation}
which is defined as the Moore-Penrose inverse of \(\bm J(\bm q)\), and $\bm J(\bm q)\bm{J}^{+}(\bm q)\hspace{-0.05cm}=\hspace{-0.05cm}\bm I_6$, where $\bm I_6$ is a $6\times 6$ identity matrix.

In addition, the null-space projection matrix is given by:
\begin{equation}
\bm{N}(\bm q) \triangleq \bm I_n - \bm{J}^{+}(\bm q)\bm{J}(\bm q) \in \Re^{n\times n},
\end{equation}
where $\bm I_n$ is an $n\times n$ identity matrix. For the definition of the null-space projection matrix, it is found that: \(\bm J(\bm q) \bm N(\bm q)\hspace{-0.05cm}=\hspace{-0.05cm}\bm 0\), \(\bm N(\bm q) \bm J^+(\bm q) =\bm 0\), and \(\bm N^2(\bm q) =\bm N(\bm q)\).

Next, the dynamic model of the robot is given by
\begin{equation}
\bm M(\bm q)\ddot{\bm q}+\bm C(\dot{\bm q}, \bm q)\dot{\bm q}+\bm g(\bm q)=\bm u+\bm J^\mathsf{T}(\bm q)\bm f_e,\label{eq_dyn}
\end{equation}
where \(\bm M(\bm q)\hspace{-0.05cm}\in\hspace{-0.05cm}\Re^{n\times n},\bm C(\dot{\bm q}, \bm q)\dot{\bm q}\hspace{-0.05cm}\in\hspace{-0.05cm}\Re^n,\bm g(\bm q)\hspace{-0.05cm}\in\hspace{-0.05cm}\Re^n\) denote the mass matrix,  Coriolis and centrifugal term, and gravity vector, respectively. \(\bm u\hspace{-0.05cm}\in\hspace{-0.05cm}\Re^n\) is the control torque, and \(\bm f_e\hspace{-0.05cm}\in\hspace{-0.05cm}\Re^6\) is the external force and torque exerted on the end effector. Note that \(\bm f_e=[f_x, f_y, f_z, \tau_x, \tau_y, \tau_z]\) is monitored such that it remains within some predefined safety range. If this external force exceeds the pre-set safety limit an emergency stop will be triggered.

For the robot physically interacting with the environment (or a human), it is common to design the control input $\bm u$ to achieve a task-space desired impedance model as follows:
\begin{equation}
\bm M_d\ddot{\tilde{\bm x}}+\bm C_d\dot{\tilde{\bm x}}+\bm K_d{\tilde{\bm x}}=\bm f_e, \label{eq_desired}
\end{equation}
which describes a dynamic relationship between the position of the robot end effector (on the left side of (\ref{eq_desired})) and the interaction force (on the right side of (\ref{eq_desired})), where \(\tilde{\bm x}\hspace{-0.05cm}=\hspace{-0.05cm}\bm x\hspace{-0.05cm}-\hspace{-0.05cm}\bm x_d\). 
The desired impedance model ensures a safe interaction with the patient, as the end effector is allowed to deviate from its trajectory to adjust to the physical contact.



\section{Multi-Modal Interaction Control}
In this section, we present a novel interaction control scheme for ultrasound scanning robots, involving the integration of and automatic transition between {\em scanning mode}, {\em recovery mode}, and {\em human-guided mode}.

\subsection{3D Vision}
\label{sec_3d_vision}

\begin{figure}[!tb]
    \centering
    \includegraphics[width=\linewidth]{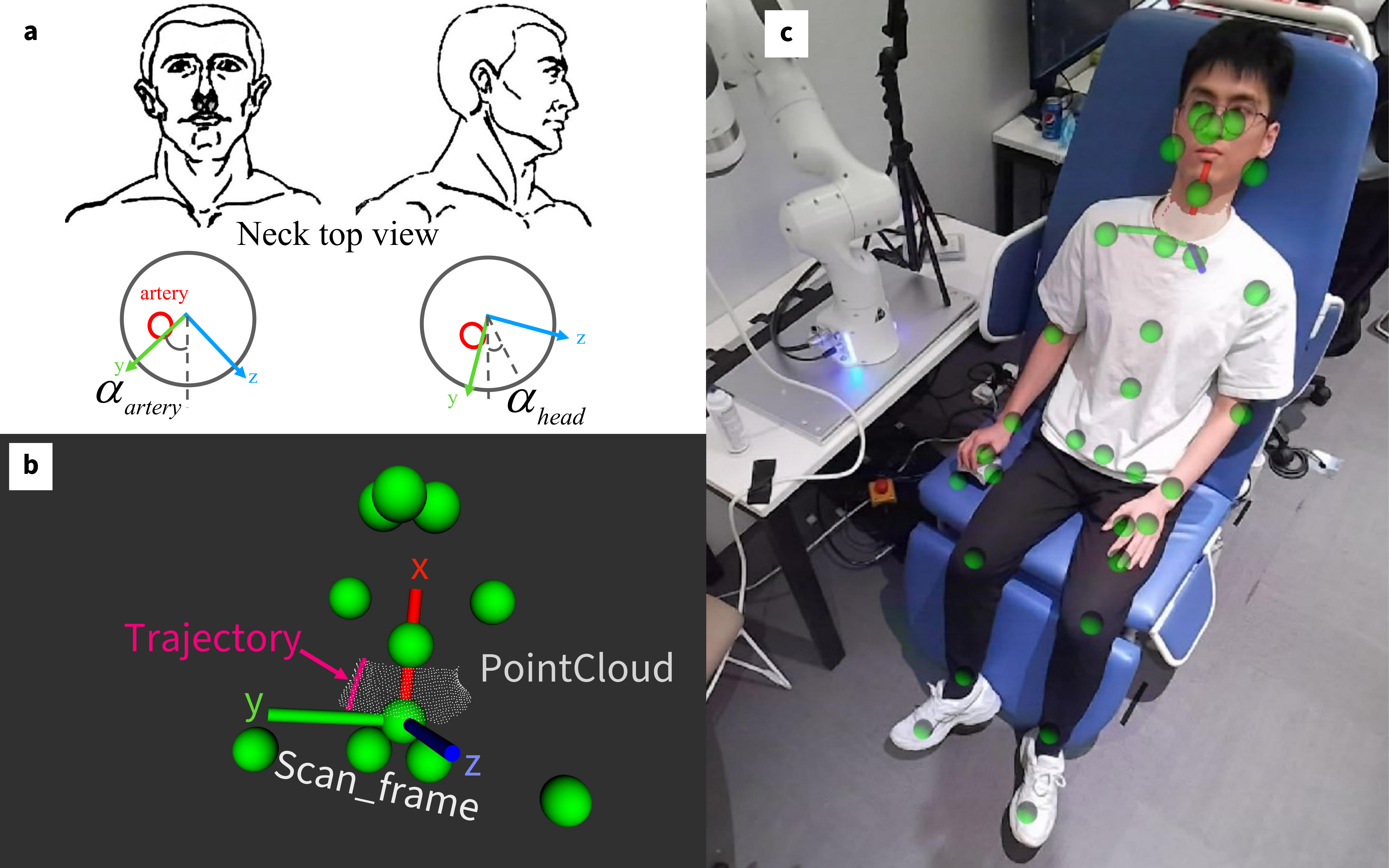}
    \caption{Diagram of the perception system. (a) The top-down view of the neck when the patient faces forward and left. (b) Scan reference frame and desired probe trajectory calculated from skeleton joints and the point cloud. (c) An RGB image of the patient, whose skeleton joints are indicated by a series of green spots.}
     \label{fig:perception-system}
\end{figure}

To accomplish the task, we need to generate the desired trajectory of the ultrasound probe and to perceive the position and motion of the patient and doctor in the environment.
Our method is based on the sensor data collected by Azure Kinect DK, which is an RGBD camera that enables body tracking and pose estimation, as shown in Fig.~\ref{fig:perception-system}(c).
The error of the depth image is less than $6 \mathrm{mm}$ at the distance of $2\mathrm{m}$\cite{tolgyessy2021,albert2020}.
The error of the neck joint is less than ${15} \mathrm{mm}$.
Therefore, the accuracy of the perception system satisfies the requirements of our autonomous system.

As a case study, this paper considers a carotid artery examination. Other scanning tasks can be performed using a similar method. 
To achieve fully autonomous ultrasound scanning of the carotid artery, a navigation system is developed to generate a real-time and patient-specific trajectory for the robot end effector.
The navigation system consists of two parts: (\romannumeral 1) a real-time scan reference frame, which follows the patient's motion, and (\romannumeral 2) a trajectory defined in the XY plane of the scan reference frame, which is calculated from the point cloud of the neck taken at a non-occlusion moment before scanning starts.\\

\noindent\textbf{Scan reference frame}:
The scan reference frame is introduced to describe the scanning task and is defined as follows: the origin is fixed at the neck joint, the x-axis points towards the head joint, representing the cervical spine, and the y-axis points towards the carotid artery (Fig.~\ref{fig:perception-system}(b)).
Therefore, a probe moving along a trajectory on the XY plane of the frame can always see the artery in its view.

The transformation from the scan reference frame to the robot base frame is defined as
\begin{equation}
    {}^{\text{base}}_{\text{scan}}\bm T = {}^{\text{base}}_{\text{probe}}\bm T \cdot 
    \begin{bmatrix}
        \bm R_x(\alpha_{\text{head}} - \alpha_{\text{artery}}) & \bm 0 \\
        \bm 0^T & 1
    \end{bmatrix},\label{tranMatrix}
\end{equation}
where \(\bm R_x(\alpha)\) is the rotation matrix describing a rotation of the x-axis, \(\alpha_{\text{head}}\) is the rotation angle of the head relative to the neck, and \(\alpha_{\text{artery}}\hspace{-0.05cm}=\hspace{-0.05cm}\pi/3\) denotes the pre-defined angle at which the y-axis of the scan reference frame points towards the artery, see Fig.~\ref{fig:perception-system}(a). Note that the second matrix on the right side of (\ref{tranMatrix}) describes the relationship between the frame of the robot end effector and the scan reference frame.\\

\noindent\textbf{Trajectory generation}: First, the point cloud is segmented by a cylindrical filter fixed to the patient’s neck to get the local point cloud of the neck. The segmented point cloud is then smoothed by the moving least squares method proposed in \cite{alexa2003}, allowing the surface of the neck to be reconstructed.
Finally, the trajectory is obtained by calculating the intersection line between the XY plane of the scan reference frame and the reconstructed surface of the neck, which is shown in Fig.~\ref{fig:perception-system}(c).
The actual desired trajectory \(\bm x_d\) is set ${3}\mathrm{cm}$ under the skin.

\subsection{Weighting Factors}
\begin{figure}[!h]
    \centering
    \includegraphics[width=\linewidth]{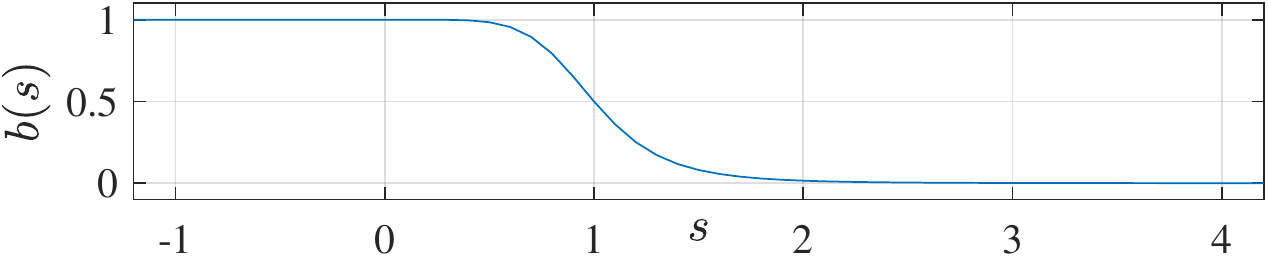}
    \caption{An illustration of \(b(s)\), the basic function of the weighting factors.}
    \label{fig:plt_ax}
\end{figure}

Three weighting factors are defined, each describing a different scenario. First of all, a basic function is defined as
\begin{subnumcases}{b(s)=}
    \frac{1}{1+s^6}\,, & \(s\geq 0\), \\
    1\,, &\(s<0\),
\end{subnumcases}
ensuring a smooth transition from $1$ to $0$ (Fig.~\ref{fig:plt_ax}), where $s$ is a general variable. All weighting factors are constructed through this basic function.

The first weighting factor $a_h$ is used to detect whether the robot end effector has been grasped,
\begin{equation}
    a_h = b\biggl(\frac{\lVert \bm d_h \rVert}{r_h}\biggr),
\end{equation}
where \(\bm d_h\) is the distance vector from the probe to one's hand and \(r_h\) is a fixed parameter.

The second weighting factor is designed to determine whether the probe is close to the patient's neck, based on a cylindrical region and defined as follows:
\begin{equation}
    a_p = b\Biggl(\frac{\sqrt{d_{py}^2+d_{pz}^2}}{r_p}\Biggr)\cdot b \Biggl( \frac{\lvert d_{px} - \frac{x_{\text{top}}+x_{\text{bottom}}}{2} \rvert}{\frac{x_{\text{top}}-x_{\text{bottom}}}{2}} \Biggr),
\end{equation}
where \(\bm d_p = [d_{px}, d_{py}, d_{pz}]^T\) is the position vector of the probe in the scan reference frame, \(r_p\) is the radius of the region, and \(x_{\text{top}}, x_{\text{bottom}}\) are the x coordinates of the top and bottom surfaces of the region, respectively.

In addition, the third weighting factor is incorporated to monitor the contact state between the scanning probe and the patient limb, which is defined as follows:
\begin{equation}
    a_f = 1 - b\biggl(\frac{^{E}f_z}{f_0}\biggr),
\end{equation}
where $f_0$ is a threshold constant, and $^Ef_z$ denotes the force along the $z$-axis in the end effector frame $\{E\}$. It is important to maintain a steady contact throughout the ultrasound scanning task such that $^Ef_z\geq0$. 

The variation of weighting factors is summarized and interpreted in Table~\ref{table_weighting_factors}, which will be used to connect the multiple interaction modes.
Note that all the weighting factors are continuous without hard-switching, and hence the transition between interaction modes is also continuous, which is important to ensure the safety of humans.
\begin{table}[!htb]
\centering
\caption{Weighting factors}
\label{table_weighting_factors}
\begin{tabular}{ccc}
\toprule
      & 0                          & 1                                  \\ \midrule
$a_h$ & probe is not grasped              & probe is grasped                   \\
$a_p$ & probe is away              & probe is close to patient \\
$a_f$ & probe doesn't contact patient & contacted            \\ \bottomrule
\end{tabular}
\end{table}

\subsection{Recovery Planning}
When the probe is far away from the patient's neck, it is risky to track the scanning trajectory purely with the impedance controller because of the potential jerking movement of the probe. Thus, a minimum-jerk trajectory generation method is applied to realize a soft recontact. 

First, a phase variable is defined as
\begin{equation}
    \phi(t)=10\biggl(\frac{t-t_i}{T_{\text{rec}}}\biggr)^3-15\biggl(\frac{t-t_i}{T_{\text{rec}}}\biggr)^4+6\biggl(\frac{t-t_i}{T_{\text{rec}}}\biggr)^5,
\end{equation}
where $t_i$ and $T$ represent the start time and duration of the current recovery behavior, and $\phi(t)\in[0,1]$. 

Next, the desired motion profile is given by
\begin{align}
    \bm{x}_d(t) &= \bm{x}_i + \phi(t)(\bm{x}_f(t)-\bm{x}_i),\label{position_profile}\\
    \dot{\bm{x}}_d(t) &= \dot{\phi}(t)(\bm{x}_f(t)-\bm{x}_i), \label{velocity_profile}
\end{align}
where $\bm{x}_f(t)$ represents the initial point on the time-varying scanning trajectory, and $\bm{x}_i$ is the end effector pose when the recovery mode is just triggered. Note that in (\ref{position_profile}) and (\ref{velocity_profile}), the quaternion part of $\bm{x}_d(t)$ is computed using spherical linear interpolation (SLERP), while the angular velocity part of $\dot{\bm{x}}_d(t)$ is set to zero due to relatively subtle changes in the desired orientation. 
Note that the planned trajectory drives the end effector into contact with the patient's body. Following that, the system transitions to the {\em scanning mode} and the probe follows the scanning trajectory planned in Section~\ref{sec_3d_vision}.

\begin{table*}[htbp]
\centering
\caption{Mode Variation}
\label{table_mode}
\begin{tabular}{ccccc}
\toprule
                  & \(a_p\) & \(a_h\) & \(a_f\) & Modification on the unified impedance model (\ref{unifiedModel}) \\ \midrule
{\em Recovery mode}     & *       & 0       & 0       & \begin{tabular}[c]{@{}c@{}c@{}} \(\bm x_d\) is given by a minimum jerk trajectory generator;\\ \(\bm q_{dn}\) is updated to current value every 100 control cycles; \(\bm K_d = \bm K_g, \bm K_{dn} = \bm K_{gn}.\)\end{tabular} \\ \hline
{\em Human-guided mode} & *       & 1       & *       & \(\bm x_d, \bm q_{dn}\) are set as the current pose, i.e., \(\bm x_d(t) = \bm x(t), \bm q_{dn}(t) = \bm q(t)\); \(\bm K_d = \bm 0, \bm K_{dn} = \bm 0.\) \\ \hline
{\em Scanning mode}     & 1       & 0       & 1       & \(\bm x_d, \bm q_{dn}\) is given by the scanning trajectory generator in section~\ref{sec_3d_vision}; \(\bm K_d = \bm K_g, \bm K_{dn} = \bm K_{gn}.\) \\ \hline
{\em Waiting mode}      & 0       & 0       & 0       & \(\bm x_d, \bm q_{dn}\) is fixed to the pose at the moment entering this mode;  \(\bm K_d = \bm K_g, \bm K_{dn} = \bm K_{gn}.\) \\
\bottomrule
\end{tabular}
\end{table*}

\subsection{Controller Development}
The controller is formulated in a classic resolved acceleration control manner \cite{1102367}, followed by a last term to decouple the external force term in the desired impedance model:
\begin{align}
    \bm u &= \bm M(\bm q)\ddot{\bm q}_c+\bm C(\dot{\bm q}, \bm q)\dot{\bm q}+\bm g(\bm q)-\bm J^\mathsf{T}(\bm q)\bm f_e\nonumber\\
    &\phantom{{}={}}\negmedspace +\bm{M}(\bm q)\bm J^{+}(\bm q)\bm{M}_d^{-1}\bm f_e,\label{eq_controller}
\end{align}
where \(\ddot{\bm q}_c\) is the command joint acceleration, which varies smoothly between interaction modes. 

Specifically,
the command joint acceleration is defined as
\begin{align}
\ddot{\bm{q}}_c &= \bm{J}^+(\bm q)[\ddot{\bm x}_d - \bm{M}_d^{-1}(\bm{C}_d \dot{\tilde{\bm{x}}}+\bm{K}_d(t) \tilde{\bm{x}})-\dot{\bm{J}}(\bm q)\dot{\bm{q}}]\nonumber\\
&\phantom{{}={}}\negmedspace +\bm{N}(\bm q)(-\bm K_{vn} \dot{\bm q}-\bm K_{dn}(t)\tilde{\bm q}),\label{eq_unified}
\end{align}
where \(\bm \tilde{\bm q}\hspace{-0.05cm}=\hspace{-0.05cm}\bm q\hspace{-0.05cm}-\hspace{-0.05cm}\bm q_{dn}\) is the joint error, \(\bm q_{dn}\in \Re^{n}\) is the desired joint position, the subscript \(n\) denotes the variable used in null space, and \(\bm{K}_d(t),\bm K_{dn}(t)\) are the stiffness parameters, which are updated according to
\begin{align}
    \bm K_d(t)&= (1-a_h)(1-a_f) \bm K_g,\label{eq_update1}\\
    \bm K_{dn}(t)&= (1-a_h)(1-a_f) \bm K_{gn}\label{eq_update2},
\end{align}
where \(\bm K_g, \bm K_{gn}\) are the maximum stiffness matrices for task space and null space, respectively, and $\bm C_d$ and $\bm K_{vn}$ are designed such that critical damping is achieved.

Substituting (\ref{eq_controller}) into (\ref{eq_dyn}), we get the closed-loop function     
\begin{equation}
\ddot{\bm q} = \ddot{\bm q}_c + \bm{J}^+(\bm q)\bm{M}_d^{-1}\bm f_e.\label{closed}
\end{equation}

Then, substituting (\ref{eq_unified}) into (\ref{closed}) yields
\begin{align}
\ddot{\bm{q}} &= \bm{J}^+(\bm q)[\ddot{\bm x}_d - \bm{M}_d^{-1}(\bm{C}_d \dot{\tilde{\bm{x}}}+\bm{K}_d(t) \tilde{\bm{x}})-\dot{\bm{J}}(\bm q)\dot{\bm{q}}]\nonumber\\
&\phantom{{}={}}\negmedspace +\bm{N}(\bm q)(-\bm K_{vn} \dot{\bm q}-\bm K_{dn}(t)\tilde{\bm q})+\bm{J}^+(\bm q)\bm{M}_d^{-1}\bm f_e. \label{eq_unified_1}
\end{align}

Multiplying both sides of (\ref{eq_unified_1}) with $\bm J(\bm q)$ and using (\ref{velEqn2}), a unified impedance model is obtained in task space as
\begin{equation}
\bm M_d\ddot{\tilde{\bm x}}+\bm C_d\dot{\tilde{\bm x}}+\bm K_d(t){\tilde{\bm x}}=\bm f_e.\label{unifiedModel}
\end{equation}
Compared with the conventional model (\ref{eq_desired}), the varying desired stiffness in (\ref{unifiedModel}) drives the robot to exhibit multiple interaction modes, according to the variation of the weighting factors. 

Specifically, when $a_p\hspace{-0.05cm}=\hspace{-0.05cm}1$ and $a_f\hspace{-0.05cm}=\hspace{-0.05cm}1$, it reduces to (\ref{eq_desired}), such that the robot tracks the reference trajectory to carry out the scanning task under a constant impedance model. When $a_h\hspace{-0.05cm}=\hspace{-0.05cm}1$, (\ref{unifiedModel}) becomes 
\begin{equation}
\bm M_d\ddot{\bm x}+\bm C_d\dot{\bm x}=\bm f_e,\label{eq_desired2}
\end{equation}
which allows the robot to follow the doctor's guidance passively. When both $a_h$ and $a_f$ reduce to zero, the robot enters the {\em recovery mode} to recontact the patient's body. The variation in weighting factors is summarized in Table~\ref{table_mode}, where we also add a {\em waiting mode}, in which the robot performs set-point impedance control when the patient and the doctor are not ready. 




Next, multiplying (\ref{closed}) with $\bm N(\bm q)$ and using the properties of the null-space projection matrix, we have
\begin{equation}
\bm N(\bm q)(\ddot{\bm q}+\bm K_{vn}\dot{\bm q}+\bm K_{dn}(t)\tilde{\bm q})=\bm 0,
\end{equation}
which describes another impedance model in null space, where \(\tilde{\bm q}\hspace{-0.05cm}=\hspace{-0.05cm}\bm q-\bm q_d\) and $\bm q_d$ represents the desired joint angles for redundant joints. The role of the desired impedance model in null space is to deal with the redundancy, that is, to quickly stabilize the joint angles at the nearest desired configuration and thus achieve the desired impedance model in task space.


\subsection{Whole Scanning Task Pipeline}
The whole scanning task pipeline is presented in Algorithm~\ref{alg_1}. The progress of the task is indicated by a variable \(t_p\hspace{-0.05cm}\in\hspace{-0.05cm}[0,T]\), where $T$ is the user-defined task time. The human-guided mode is the highest priority once contact between a human hand and the probe is detected, as shown in line~\ref{alg_guiding_if}. Here, the subscript $t$ means threshold, the same as below. The trajectory in line~\ref{alg_waiting_if},~\ref{alg_traj_if} is obtained through the method in section~\ref{sec_3d_vision}, which can alternatively be replaced by any high-level navigation/planning system, showing the portability of our interaction control method. In line~\ref{eq_tp}, the progress time increments only in the scanning mode, guaranteeing that every point in the trajectory is traversed. Specifically, at progress time $t_p$, the desired pose \(\bm x_d(t_p)=\bm x_d(i)\). The right term denotes the \(i\)th trajectory point, where
\begin{equation}
    i=\biggl[\frac{t_p\times N}{T}\biggr].\label{eq_calc_desired}
\end{equation}

Note that the transition between scanning mode and recovery mode is dependent on the error, weighting factors $a_p$ and $a_f$ (line~\ref{alg_traj_error_if}), that is, a position-force combined logic operation, which is useful for dealing with frequent physical interaction in the doctor-robot co-existing scenario. 
\begin{algorithm}[!t]
\caption{Multi-Modal Interaction Control Algorithm}\label{alg_1}
\begin{algorithmic}[1]
\State progress time \(t_p=0\)
\While {$t_p < T$}
    \If{$a_h \ge a_{ht}$}  \label{alg_guiding_if}
        \State {\em Human-guided mode};
    \ElsIf{no trajectory} \label{alg_waiting_if}
        \State {\em Waiting mode};
    \ElsIf{have a trajectory $\{\bm x_d(i)\}_{i=1}^N$}\label{alg_traj_if}
        \State calculate the desired pose $\bm x_d(t_p)$ by (\ref{eq_calc_desired});
        \If {$(\lVert \tilde{\bm x}  \rVert<\epsilon)$ and $(a_p>a_{pt})$ or $(a_f>a_{ft})$}\label{alg_traj_error_if}
            \State {\em Scanning mode};
            \State $t_p = t_p+dt$;\label{eq_tp}
        \Else 
            \State {\em Recovery mode};
        \EndIf
    \EndIf
\EndWhile
\end{algorithmic}
\end{algorithm}

\begin{figure} [!tb]
    \centering
    \includegraphics[width=\linewidth]{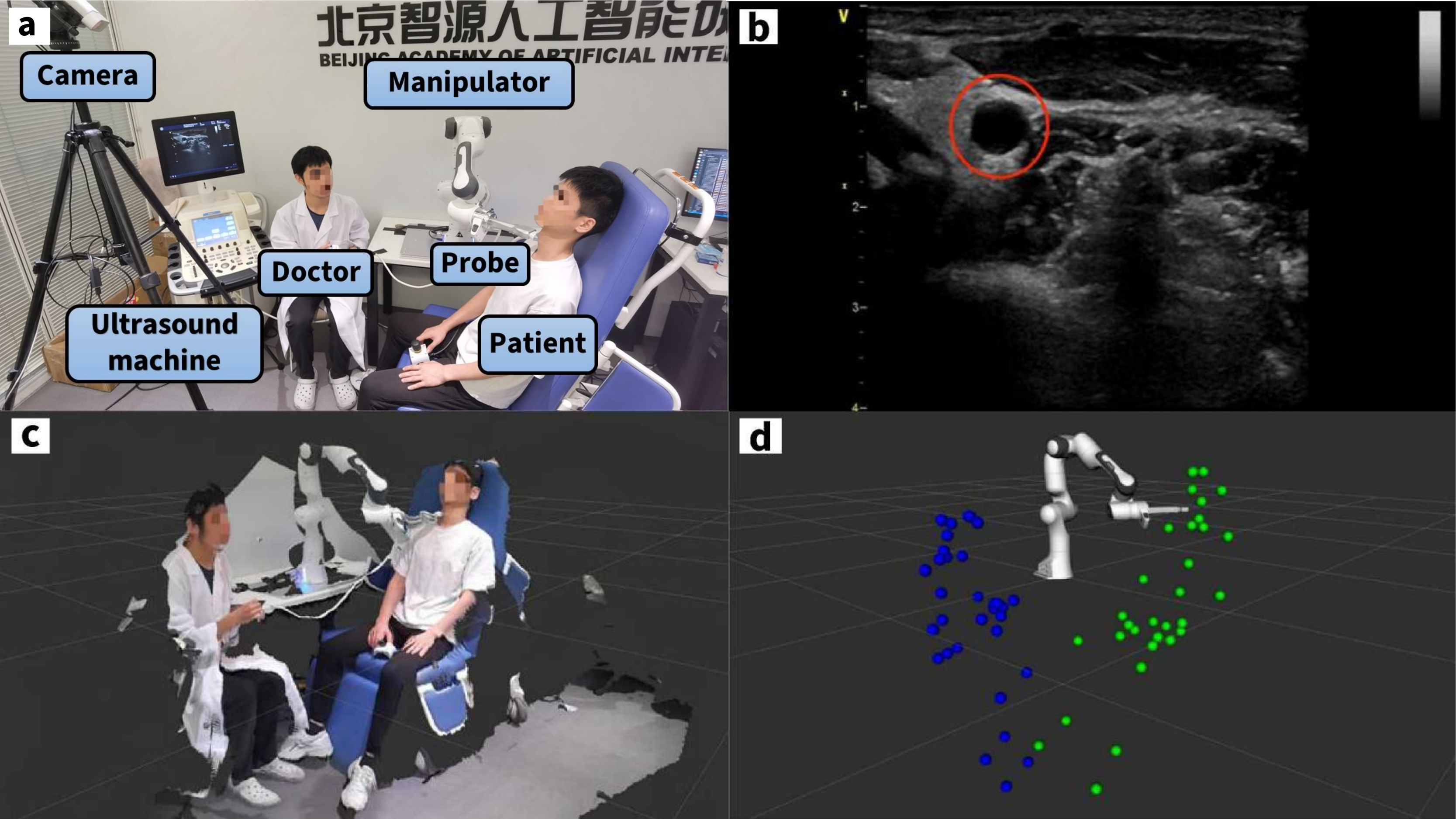}
    \caption{The experimental setup: (a) The overall system consists of an ultrasound machine, a RGBD camera, a manipulator, an ultrasound probe, and PCs; (b) An ultrasound image, where the red circle indicates the position of the image feature, that is, the cross-section carotid artery; (c) A point-cloud image generated by Azure Kinect DK; (d) The corresponding skeleton points of the doctor (left) and the patient (right). Note that the four images were taken at the same time.}
  \label{fig:exp_setup}
\end{figure}

\section{Experiment}
Experiments were carried out to validate the proposed control method. The experimental setup is shown in  Fig.~\ref{fig:exp_setup}.
The overall system consisted of five parts: (i) an ultrasound system (Vivid E7), including an imaging machine and a probe; (ii) a 7-DOF robot manipulator (Franka Panda); (iii) an RGBD camera (Azure Kinect DK); (iv) a hospital bed (which can be folded as a chair); and (v) two PCs for processing the camera data and controlling the robot, respectively.

\begin{figure}[!tb]
\centering
\includegraphics[width=1.8in]{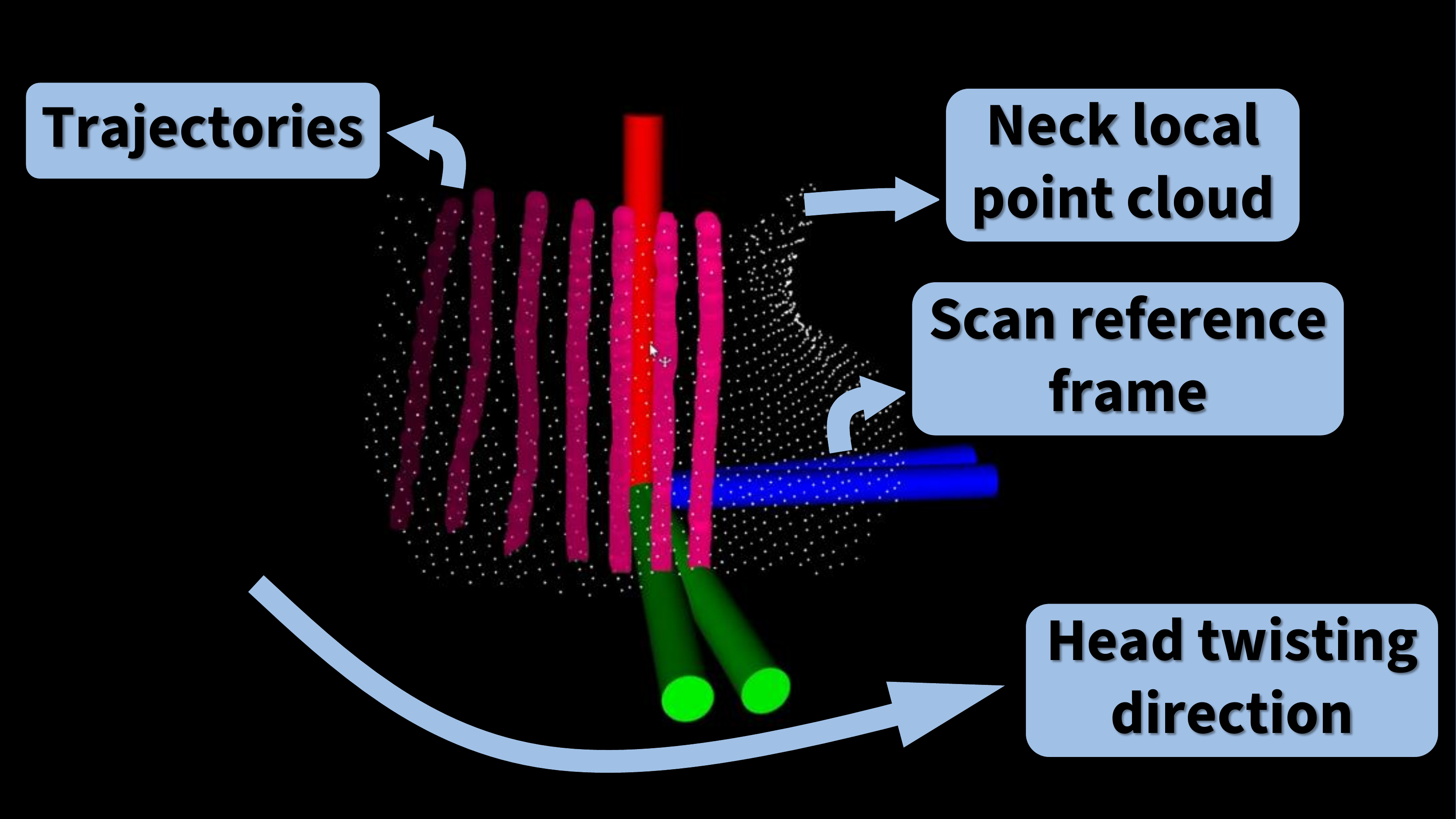}
\caption{A series of scanning trajectories generated with the neck's local point cloud. Each trajectory corresponds to a different instance in time as the patient turns head.}\label{fig_traj_pintu}
\end{figure}

In the experiments, a human subject (the patient) is sat on the hospital bed while the robot moves the scanning probe along the surface of his neck to perform the ultrasound imaging. A doctor is sat beside the patient, prepared to intervene and manually guide the motion of the probe during the scanning task. The generated scanning trajectories on the patient's neck are shown in 
Fig.~\ref{fig_traj_pintu}. 

To simulate the complexity of the scanning task, we considered a restless patient who would disturb the task by dodging the probe, moving his body, sneezing during the scanning, and pushing the probe deliberately. The doctor would then intervene to handle some unexpected situations. Hence, the experiment demonstrated the ability of the robot's control system to transition between multiple interaction modes in order to adapt to different situations, according to the control parameters shown in Table~\ref{task2_params}.
\begin{table}[!th]
\centering
\caption{Control parameters in the experiment}
\label{task2_params}
\begin{tabular}{cccccc}
\toprule
\textbf{Parameter}  & \textbf{Value}      & \textbf{Parameter} & \textbf{Value}     & \textbf{Parameter} & \textbf{Value} \\ \midrule
$r_h$               & \({0.2}\mathrm{m}\)   & $r_p$              & \({0.15}\mathrm{m}\) & $a_{ht}$           & 0.8            \\
$x_{\text{top}}$    & \({0.1}\mathrm{m}\)   & $f_{z}$            & \({12.5}\mathrm{N}\) & $a_{ft}$           & 0.5            \\
$x_{\text{bottom}}$ & \({-0.02}\mathrm{m}\) & $T$                & \({30}\mathrm{s}\)   & $a_{pt}$           & 0.9            \\
$\epsilon$          & \({0.05}\mathrm{m}\)  & $T_{\text{rec}}$   & \({8.0}\mathrm{s}\)  & $N$                & 100            \\ \bottomrule
\end{tabular}
\end{table}
\begin{figure} [!t]
  \centering
    \includegraphics[width=\linewidth]{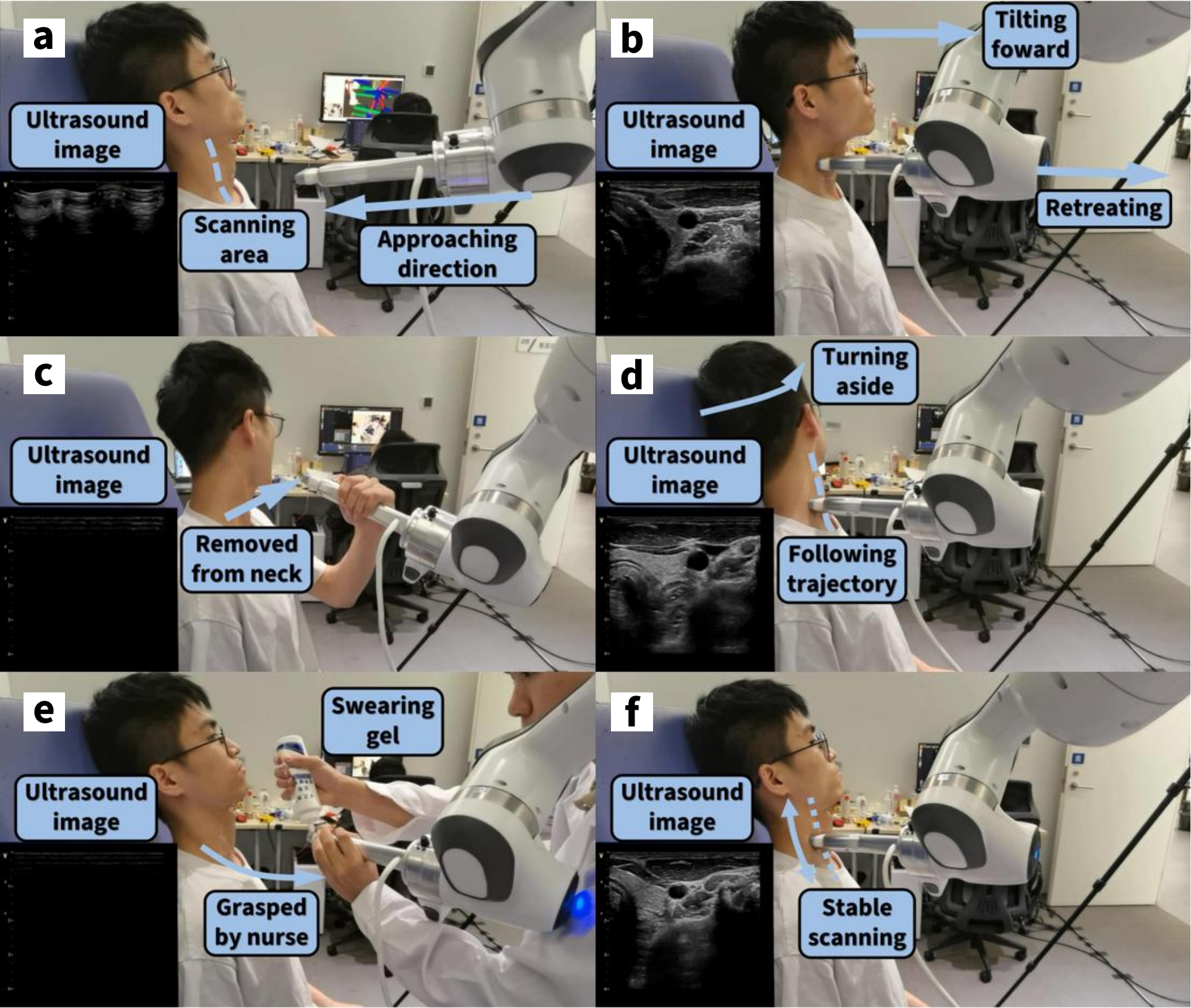}
  \caption{Snapshots at different times: (a) The probe was approaching the desired scanning area in the {\em recovery mode}; (b) The probe retreated while the patient tilted forward; (c) The probe was temporarily removed from the patient's neck; (d) The probe followed the desired scanning trajectory while the patient turned to the side; (e) The probe was held by the doctor for coupling gel to be applied; (f) The probe was operating in the {\em scanning mode}.}
  \label{task2_snapshots}
\end{figure}
\begin{figure}[!t]
    \includegraphics[width=0.95\linewidth]{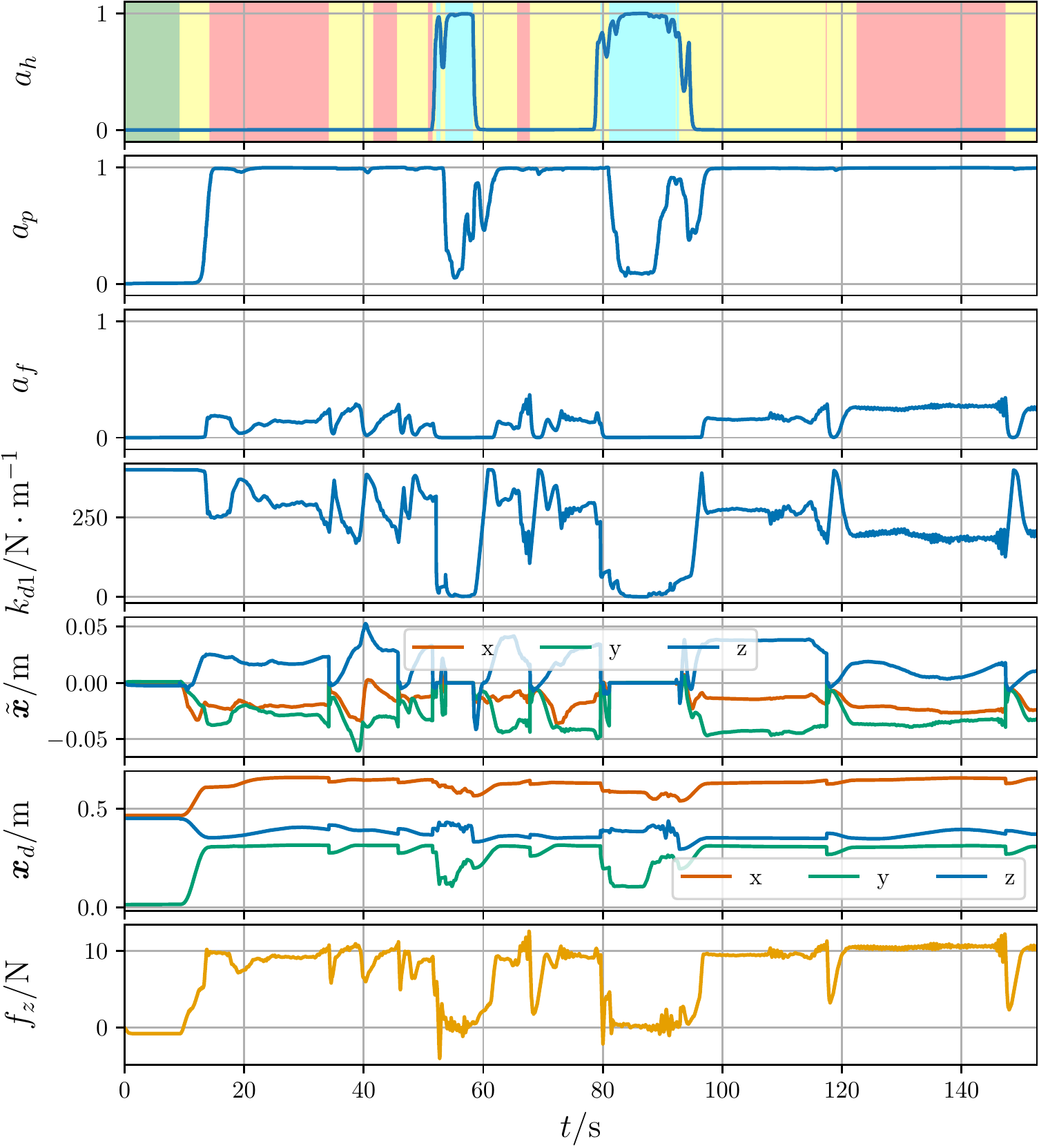}
 \caption{Experimental results, from top to bottom: the first three subfigures correspond to the weighting factors of $a_h$, $a_p$, $a_f$ respectively; the fourth subfigure plots the translational stiffness of $k_{d1}$; the fifth subfigure plots the translational tracking error of $\tilde{\bm{x}}$; the sixth subfigure plots the desired translation of $\bm{x}_d$; the seventh subfigure plots the estimated external force of $^{E}{f}_{z}$ expressed in end effector frame $\{E\}$. Background color in the first subfigure denotes different interaction modes, specifically, green - {\em waiting mode}, yellow - {\em recovery mode}, red - {\em scanning mode}, blue - {\em human-guided mode}.}
 \label{fig_task2_result}
\end{figure}

The snapshots at different times during the procedure are shown in Fig.~\ref{task2_snapshots}, and the experimental results are shown in Fig.~\ref{fig_task2_result}. The scanning task progressed as follows:

\begin{enumerate}
    \item The {\em recovery mode} was activated between \(10\)s and \(15\)s, such that $\bm{x}_d$ gradually changed to the desired value, and $a_p$ continuously increased to \(1\). As a result, the scanning probe approached the patient to establish physical contact (see Fig.~\ref{task2_snapshots}(a)). 

    \item From ${35}\mathrm{s}$ to ${45}\mathrm{s}$, the patient intentionally pushed the probe forward twice, producing a shift in $^{E}{f}_{z}$ and subsequently in $a_{f}$ as a result of the interaction forces. The combined effect of the increasing $a_{f}$ and tracking error $\tilde{\bm{x}}$ triggered the {\em recovery mode}; Then, the error dropped, and the robot subsequently returned to the {\em scanning mode} (see Fig.~\ref{task2_snapshots}(b)). Within such a period, $a_{f}$ played an important role in attenuating the amplitude of $k_{d1}$, and thus in limiting the contact force $^{E}{f}_{z}$. Therefore, the patient's safety and comfort were guaranteed. 

    \item During ${51}\mathrm{s}$ to ${59}\mathrm{s}$, the probe was manually moved away from the patient's neck (see Fig.~\ref{task2_snapshots}(c)). The translation stiffness $k_{d1}$ rapidly dropped to \(0\) along with $a_h$, indicating the transition into the {\em human-guided mode} in which the robot becomes passive and receptive to manual guidance. Accordingly, the desired translation $\bm{x}_d$ was also the probe's instantaneous position. After the probe was released, $k_{d1}$ returned to large values, to make the probe stay in its current pose. 

    \item In Fig.~\ref{task2_snapshots}(d), the patient turned his head to dodge the probe at \(68\) s. The drop in $^{E}{f}_{z}$ and the increase in $\tilde{\bm{x}}$ activated the {\em scanning mode}, making the contact stable to maintain high-quality ultrasound imaging.
    
    \item Similar to the period of 3), the robot once again entered the {\em human-guided mode} again between 79s and 95s, when the doctor intervened to apply coupling gel (see Fig.~\ref{task2_snapshots}(e)).

    \item The robot was operating in the {\em scanning mode} between 14s and 34s and from 122s to 146s, where the scanning task and ultrasound image acquisition of the patient's carotid artery were performed simultaneously. The robot maintained a stable and steady contact with the patient, such that ${a}_{p}\approx {1}{}$; In addition, the contact force was kept at an appropriate value which was sufficient for high-quality imaging while also ensuring the safety of the patient, $^{E}{f}_{z}\hspace{-0.05cm}\approx\hspace{-0.05cm}{10}\mathrm{N}$ and $a_{f}\hspace{-0.05cm}\leq\hspace{-0.05cm}{0.3}$, the tracking error was small, $\lVert\tilde{\bm{x}}\rVert\hspace{-0.05cm}\leq\hspace{-0.05cm}{0.04}\mathrm{m}$, and the ultrasound images were of high-quality, i.e., clear edge of the vascular walls. The scanning probe followed the changing desired trajectory in the presence of disturbances such as the patient moving or turning his head (see Fig.~\ref{task2_snapshots}(f)); Throughout the procedure, the image quality remained consistent, albeit with transient shadows (see the ultrasound images at the bottom left of each snapshot in Fig. \ref{task2_snapshots}).
\end{enumerate}

As shown in Fig.~\ref{fig_task2_result}, the transition between the multiple interaction modes was safe, stable, and continuous throughout the scanning procedure, demonstrating the effectiveness of the proposed control scheme. 

\section{Conclusions}
In this paper we proposed a novel muti-modal interaction control scheme for ultrasound scanning robots, which closely collaborate with a doctor in a co-existing environment. Specifically, different control modes for normal scanning, recontacting, and human-guided situations have been smoothly and stably integrated into a unified control input. Furthermore, the control scheme automatically transitions between different control modes according to the doctor’s actions and changes in the environment, such as movement of the patient's body or disturbance to the probe's trajectory. As a result, it allows the doctor to safely interact with the robot to modify its motion at any time. Such a feature is important for a scanning robot, because (i) it can well combine the advantages of the doctor's experience/knowledge and the robot's autonomous ability, and (ii) it can quickly deal with an emergency or other unforeseen scenarios with the safe involvement of doctors.  
Experimental results on a cross-section carotid examination, as an example test scenario, have been presented to validate the proposed control scheme.  The experimental results demonstrate that both the safety and the realization of the unified impedance model are achieved even in the presence of significant movement of the patient, involvement of the doctor, and other unexpected situations. 

Future work will be twofold. 1) A more systematic assessment and improvement on the ultrasound image quality is needed. 2) The usage of variable stiffness in (\ref{unifiedModel}) destroys the passivity of the control system, which could be restored by tank-based technique \cite{ferraguti2015energy}.

\bibliographystyle{ieeetr}
\bibliography{main}

\begin{thebibliography}{10}

\bibitem{shung2011diagnostic}
K.~K. Shung, ``Diagnostic ultrasound: Past, present, and future,'' {\em J Med
  Biol Eng}, vol.~31, no.~6, pp.~371--4, 2011.

\bibitem{Li2021}
K.~Li, Y.~Xu, and M.~Q.-H. Meng, ``{An overview of systems and techniques for
  autonomous robotic ultrasound acquisitions},'' {\em IEEE Transactions on
  Medical Robotics and Bionics}, vol.~3, no.~2, pp.~510--524, 2021.

\bibitem{selvaggio2021autonomy}
M.~Selvaggio, M.~Cognetti, S.~Nikolaidis, S.~Ivaldi, and B.~Siciliano,
  ``Autonomy in physical human-robot interaction: A brief survey,'' {\em IEEE
  Robotics and Automation Letters}, 2021.

\bibitem{nakadate2010implementation}
R.~Nakadate, J.~Solis, A.~Takanishi, E.~Minagawa, M.~Sugawara, and K.~Niki,
  ``Implementation of an automatic scanning and detection algorithm for the
  carotid artery by an assisted-robotic measurement system,'' in {\em 2010
  IEEE/RSJ International Conference on Intelligent Robots and Systems},
  pp.~313--318, IEEE, 2010.

\bibitem{hennersperger2016towards}
C.~Hennersperger, B.~Fuerst, S.~Virga, O.~Zettinig, B.~Frisch, T.~Neff, and
  N.~Navab, ``Towards mri-based autonomous robotic us acquisitions: a first
  feasibility study,'' {\em IEEE transactions on medical imaging}, vol.~36,
  no.~2, pp.~538--548, 2016.

\bibitem{kojcev2016dual}
R.~Kojcev, B.~Fuerst, O.~Zettinig, J.~Fotouhi, S.~C. Lee, B.~Frisch, R.~Taylor,
  E.~Sinibaldi, and N.~Navab, ``Dual-robot ultrasound-guided needle placement:
  closing the planning-imaging-action loop,'' {\em International journal of
  computer assisted radiology and surgery}, vol.~11, no.~6, pp.~1173--1181,
  2016.

\bibitem{8002599}
P.~Chatelain, A.~Krupa, and N.~Navab, ``Confidence-driven control of an
  ultrasound probe,'' {\em IEEE Transactions on Robotics}, vol.~33, no.~6,
  pp.~1410--1424, 2017.

\bibitem{huang2018fully}
Q.~Huang, B.~Wu, J.~Lan, and X.~Li, ``Fully automatic three-dimensional
  ultrasound imaging based on conventional b-scan,'' {\em IEEE transactions on
  biomedical circuits and systems}, vol.~12, no.~2, pp.~426--436, 2018.

\bibitem{welleweerd2021out}
M.~K. Welleweerd, A.~G. de~Groot, V.~Groenhuis, F.~J. Siepel, and
  S.~Stramigioli, ``Out-of-plane corrections for autonomous robotic breast
  ultrasound acquisitions,'' in {\em 2021 IEEE International Conference on
  Robotics and Automation (ICRA)}, pp.~12515--12521, IEEE, 2021.

\bibitem{tan2022flexible}
J.~Tan, B.~Li, Y.~Li, B.~Li, X.~Chen, J.~Wu, B.~Luo, Y.~Leng, Y.~Rong, and
  C.~Fu, ``A flexible and fully autonomous breast ultrasound scanning system,''
  {\em IEEE Transactions on Automation Science and Engineering}, 2022.

\bibitem{li2021image}
K.~Li, Y.~Xu, J.~Wang, D.~Ni, L.~Liu, and M.~Q.-H. Meng, ``Image-guided
  navigation of a robotic ultrasound probe for autonomous spinal sonography
  using a shadow-aware dual-agent framework,'' {\em IEEE Transactions on
  Medical Robotics and Bionics}, vol.~4, no.~1, pp.~130--144, 2021.

\bibitem{bi2022vesnet}
Y.~Bi, Z.~Jiang, Y.~Gao, T.~Wendler, A.~Karlas, and N.~Navab, ``Vesnet-rl:
  Simulation-based reinforcement learning for real-world us probe navigation,''
  {\em IEEE Robotics and Automation Letters}, 2022.

\bibitem{amanhoud2020force}
W.~Amanhoud, M.~Khoramshahi, M.~Bonnesoeur, and A.~Billard, ``Force adaptation
  in contact tasks with dynamical systems,'' in {\em 2020 IEEE International
  Conference on Robotics and Automation (ICRA)}, pp.~6841--6847, IEEE, 2020.

\bibitem{kronander2015passive}
K.~Kronander and A.~Billard, ``Passive interaction control with dynamical
  systems,'' {\em IEEE Robotics and Automation Letters}, vol.~1, no.~1,
  pp.~106--113, 2015.

\bibitem{schindlbeck2015unified}
C.~Schindlbeck and S.~Haddadin, ``Unified passivity-based cartesian
  force/impedance control for rigid and flexible joint robots via task-energy
  tanks,'' in {\em 2015 IEEE international conference on robotics and
  automation (ICRA)}, pp.~440--447, IEEE, 2015.

\bibitem{dyck2022impedance}
M.~Dyck, A.~Sachtler, J.~Klodmann, and A.~Albu-Sch{\"a}ffer, ``Impedance
  control on arbitrary surfaces for ultrasound scanning using discrete
  differential geometry,'' {\em IEEE Robotics and Automation Letters}, vol.~7,
  no.~3, pp.~7738--7746, 2022.

\bibitem{dietrich_overview_2015}
A.~Dietrich, C.~Ott, and A.~Albu-Schäffer, ``An overview of null space
  projections for redundant, torque-controlled robots,'' {\em The International
  Journal of Robotics Research}, vol.~34, no.~11, pp.~1385--1400, 2015.
\newblock \_eprint: https://doi.org/10.1177/0278364914566516.

\bibitem{sadeghian2011multi}
H.~Sadeghian, L.~Villani, M.~Keshmiri, and B.~Siciliano, ``Multi-priority
  control in redundant robotic systems,'' in {\em 2011 IEEE/RSJ International
  Conference on Intelligent Robots and Systems}, pp.~3752--3757, IEEE, 2011.

\bibitem{oh1998extended}
Y.~Oh, W.~Chung, and Y.~Youm, ``Extended impedance control of redundant
  manipulators based on weighted decomposition of joint space,'' {\em Journal
  of Robotic Systems}, vol.~15, no.~5, pp.~231--258, 1998.

\bibitem{sadeghian2013task}
H.~Sadeghian, L.~Villani, M.~Keshmiri, and B.~Siciliano, ``Task-space control
  of robot manipulators with null-space compliance,'' {\em IEEE Transactions on
  Robotics}, vol.~30, no.~2, pp.~493--506, 2013.

\bibitem{tolgyessy2021}
M.~T{\"o}lgyessy, M.~Dekan, {\v L}.~Chovanec, and P.~Hubinsk{\'y}, ``Evaluation
  of the {{Azure Kinect}} and {{Its Comparison}} to {{Kinect V1}} and {{Kinect
  V2}},'' {\em Sensors}, vol.~21, p.~413, Jan. 2021.

\bibitem{albert2020}
J.~A. Albert, V.~Owolabi, A.~Gebel, C.~M. Brahms, U.~Granacher, and B.~Arnrich,
  ``Evaluation of the {{Pose Tracking Performance}} of the {{Azure Kinect}} and
  {{Kinect}} v2 for {{Gait Analysis}} in {{Comparison}} with a {{Gold
  Standard}}: {{A Pilot Study}},'' {\em Sensors}, vol.~20, p.~5104, Sept. 2020.

\bibitem{alexa2003}
M.~Alexa, J.~Behr, D.~{Cohen-Or}, S.~Fleishman, D.~Levin, and C.~Silva,
  ``Computing and rendering point set surfaces,'' {\em IEEE Transactions on
  Visualization and Computer Graphics}, vol.~9, pp.~3--15, Jan. 2003.

\bibitem{1102367}
J.~Luh, M.~Walker, and R.~Paul, ``Resolved-acceleration control of mechanical
  manipulators,'' {\em IEEE Transactions on Automatic Control}, vol.~25, no.~3,
  pp.~468--474, 1980.

\bibitem{ferraguti2015energy}
F.~Ferraguti, N.~Preda, A.~Manurung, M.~Bonfe, O.~Lambercy, R.~Gassert,
  R.~Muradore, P.~Fiorini, and C.~Secchi, ``An {{Energy Tank-Based Interactive
  Control Architecture}} for {{Autonomous}} and {{Teleoperated Robotic
  Surgery}},'' {\em IEEE Transactions on Robotics}, vol.~31, no.~5,
  pp.~1073--1088, 2015.

\end{thebibliography}

\end{document}